\documentclass[letterpaper, 10 pt, conference]{ieeeconf}  
                                                          
\IEEEoverridecommandlockouts                              
                                               
\usepackage{graphicx}
\usepackage{amsmath}
\usepackage{bm}
\usepackage{url}
\usepackage{subcaption}
\usepackage{xcolor}

\title{\LARGE \bf
Kinesthetic Teaching in Robotics: a Mixed Reality Approach
}

\author{Simone~Macci\`o, Mohamad Shaaban, Alessandro~Carf\`i, and Fulvio~Mastrogiovanni
\thanks{This work has been supported by the European Union Erasmus+ Programme via the Joint Master Degree program European Master on Advanced Robotics Plus (EMARO+), and via the Italian government support under the National Recovery and Resilience Plan (NRRP), Mission 4, Component 2 Investment 1.5, funded from the European Union NextGenerationEU and awarded by the Italian Ministry of University and Research..}
\thanks{All the authors are with the Department of Informatics, Bioengineering, Robotics, and Systems Engineering, University of Genoa, Via Opera Pia 13, 16145 Genoa, Italy (Corresponding author email: simone.maccio@edu.unige.it)}%
}

\begin{document}

\maketitle
\pagestyle{empty}

\begin{abstract}

As collaborative robots become more common in manufacturing scenarios and adopted in hybrid human-robot teams, we should develop new interaction and communication strategies to ensure smooth collaboration between agents. In this paper, we propose a novel communicative interface that uses Mixed Reality as a medium to perform Kinesthetic Teaching (KT) on any robotic platform. We evaluate our proposed approach in a user study involving multiple subjects and two different robots, comparing traditional physical KT with holographic-based KT through user experience questionnaires and task-related metrics.
\end{abstract}


\section{Introduction}
\label{section:intro}
In smart factories, robots are expected to coexist and work alongside humans rather than replace them. This new manufacturing paradigm has led to the development of collaborative robots, which are adaptive and highly versatile platforms \cite{wang2020overview} that can work alongside human workers. Despite its growing popularity, Human-Robot Collaboration (HRC) is still far from reaching maturity, as multiple research facets are yet to be tackled. One such aspect involves developing a structured communication enabling agents to exchange information intuitively \cite{suzuki2022augmented}. As multiple social studies have shown \cite{mutlu2009nonverbal, calisgan2012identifying}, effective bi-directional communication is crucial for successful collaboration, as it allows agents to infer each other's actions, synchronize, and receive appropriate feedback from their teammates. Conversely, poor communication can lead to misunderstandings, failed interactions, and consequent distrust in the robot teammate \cite{ye2019human}.

Designing a comprehensive communication interface is a complex task that requires selecting an appropriate communicative channel. One of the most promising approaches combines Mixed Reality (MR) with wearable Head-Mounted Displays (HMD), enabling the creation of engaging holographic interfaces where users perceive 3D digital content superimposed onto the surrounding scene \cite{ostanin2020human}. This virtual layer can act as a communicative channel to achieve intuitive human-robot communication. In this regard, few works have focused on using MR to preview a robot's intentions and upcoming actions \cite{rosen2019communicating, newbury2022visualizing, maccio2022mixed}, offering helpful visual feedback to the human teammate during collaboration. In our previous work \cite{maccio2022mixed}, we mainly focused on robot-to-human communication, introducing the concept of \textit{communicative act} and formalizing the communication for conveying the robot's intentions via holographic cues.

\begin{figure}[t!]
    \centering
    \includegraphics[width=0.42\textwidth]{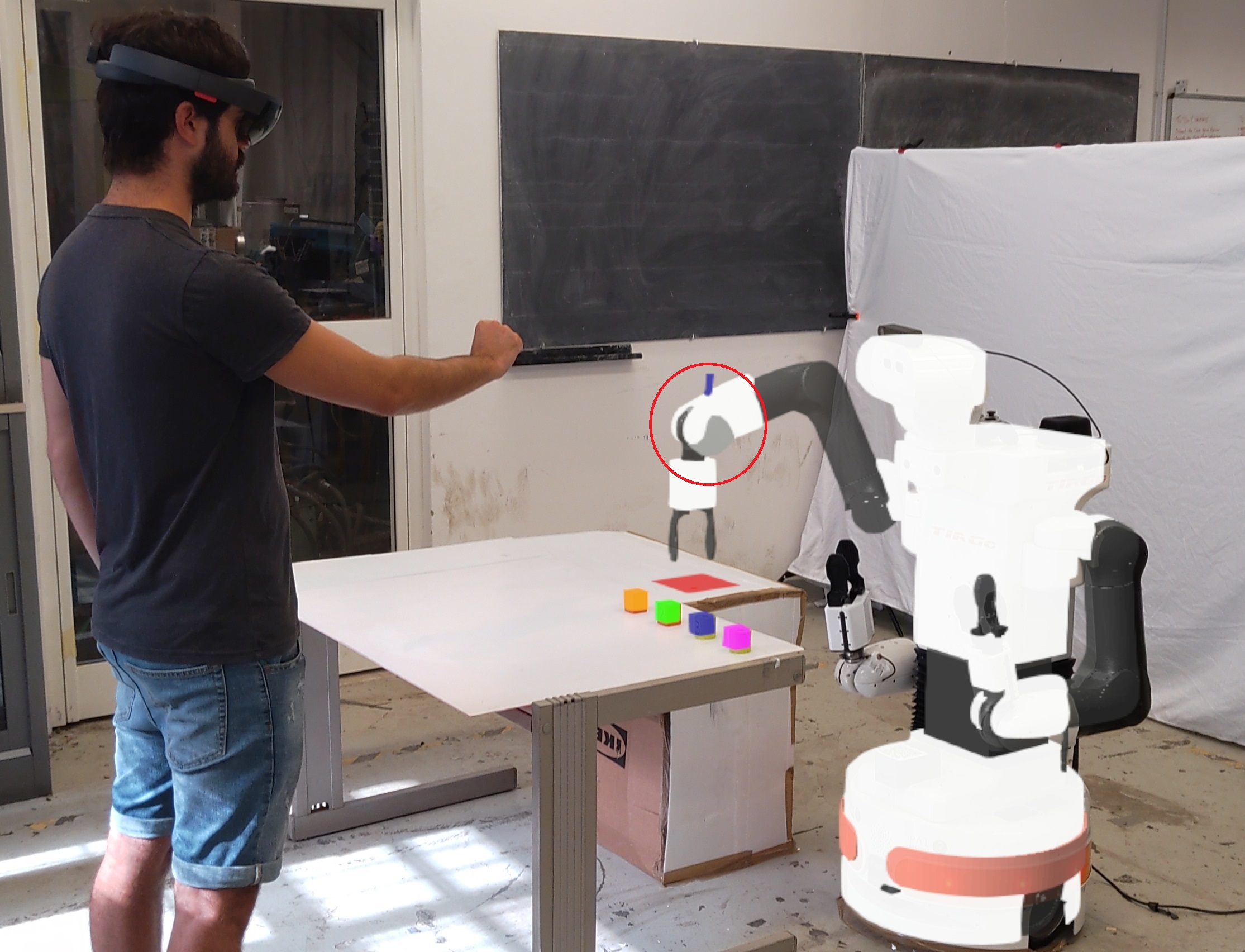}
    \caption{An experimenter in the middle of a holographic KT session with the Tiago++ robot.  By interacting with and manipulating the grey holographic sphere, superimposed on the digital robot's wrist and here highlighted via a red circle, the user can teach actions to the robot teammate using gestures and voice. }
    \label{fig:tiago-kt}
\end{figure}

In this paper, we investigate human-to-robot communication by leveraging MR to allow operators to \textit{teach} robots through holographic communication.  In particular, we embrace the Learning from Demonstration (LfD) approach \cite{ravichandar2020recent}, postulating that LfD sessions can be viewed as communication acts aimed at transferring skills from a human operator to a robot teammate through explicit actions or gestures. Specifically, our work is focused on one branch of LfD, namely Kinesthetic Teaching (KT), a well-known teaching technique in which human operators manually drive the robot's arm or end-effector, enabling the machine to learn new actions from direct demonstration. In the context of this work,  we claim that such teaching methodology can be framed into the communicative space introduced in \cite{maccio2022mixed}. Therefore, throughout the paper, we provide an analytical formalization of KT in the proposed communicative framework and translate it into a modular software component, which enables KT in human-robot interactive scenarios through holographic communication. Our proposed approach, while leveraging MR for intuitive and straightforward communication between humans and robots,  adheres to the LfD paradigm, providing a holographic tool to demonstrate skills to the robot teammate in HRC. Furthermore, given the unconstrained nature of the MR space where the KT session takes place, our proposed strategy potentially opens up the possibility of performing KT on any robotic platform compatible with the Universal Robot Description Format (URDF).

In addition to presenting such a holographic-based tool for KT, we evaluate its effectiveness in demonstrating tasks to robots and its perceived user experience (UX). Specifically, we claim that the holographic-based KT approach can serve as a suitable alternative to traditional, hand-guided KT in scenarios where the latter is not available or not implemented for a particular robot platform. To test this hypothesis, we conducted a preliminary user study with 12 subjects and two robots, comparing the two KT alternatives using task-based metrics and UX questionnaires.

The paper is organized as follows. Section \ref{section:background} reports a review of relevant literature. Section \ref{section:formalization} formalizes KT inside the holographic communication space, whereas Section \ref{section:softarchitecture} details the implementation of the software components. Section \ref{section:experimental} and Section \ref{scetion::results} respectively discuss the experimental scenario devised to test the holographic KT approach and the user study results. Finally, Section \ref{section:conclusions} provides conclusions and possible extensions for this work.

\section{Background}
\label{section:background}

Over the years, various communication strategies have been explored and adopted in HRC, involving both explicit media (e.g., voice \cite{van2012pick}, upper limb gestures \cite{carfi2018online, bongiovanni2022gestural}, light and visual cues \cite{cha2016using, song2018bioluminescence}) and implicit ones (e.g. gaze \cite{kalegina2018characterizing}, posture and body motions \cite{s20216347}). However, most of these approaches have intrinsic limitations and cannot be employed for developing a bi-directional communication interface, thus limiting their adoption to a subset of collaborative applications. For example, human-like communication involving gestures and gaze may be expressive and intuitive, but most collaborative platforms physically lack the features needed to replicate such cues. 

With the introduction of Augmented Reality (AR) in mobile devices like smartphones and tablets, a new virtual layer could be exploited by researchers to enable intuitive and straightforward communication between human and robot teammates \cite{michalos2016augmented, chacko2019augmented, chandan2021arroch}. This approach has become even more relevant with the adoption of MR-HMD devices, which offer a whole new level of immersion and make it possible to develop interfaces for either programming robots' behaviours \cite{quintero2018robot, wang2020closed, chan2022design} or getting intuitive feedback throughout the interaction \cite{williams2019mixed, rosen2020mixed}. In this context, researchers also focused on conveying robot's intentions via MR, evaluating intuitive and expressive strategies for robots to anticipate their actions via holographic cues during interactive tasks efficiently \cite{rosen2019communicating, newbury2022visualizing, maccio2022mixed}. 

While extensive research covers how robots can effectively communicate with human teammates via MR, only a few works have explored how we can leverage this holographic medium for intuitive and straightforward human-to-robot communication,  particularly in LfD. In this context, popular approaches at LfD rely on computer vision to transfer desired motions using passive observation of human actions \cite{qiu2020hand, eze2024learning}, or make use of hand-tracking devices to teach skills through teleoperation-based LfD \cite{si2021review}. While providing a straightforward communication interface to transfer skills to the robotic teammate, these approaches generally require a structured environment and complex calibration routines, which may limit their application in real-world settings. On the contrary, adopting MR as a communication medium for LfD could mitigate these drawbacks, as MR-HMDs are naturally designed for unstructured environments and could provide similar demonstration capabilities with minimum calibration and setup.

Focusing on the particular branch of KT, some of the earliest attempts at combining KT and MR  still relied on the physical robot for hand guidance and demonstration and employed the holographic medium only for later visualizing the learned robot action and for adding constraints to the motion \cite{luebbers2019augmented, luebbers2021arc}. MR-based communication to achieve KT is foreshadowed in \cite{8967649}, where the authors exploit the hand-tracking capabilities of MR-HMD devices to manually drive the individual joints of an industrial robotic manipulator, teaching motions to the machine in the process.  Similarly, in \cite{pinto2020multimodal} a system is presented where a tabletop holographic robot can be taught a simple pick-and-place task via holographic hand guidance. Finally, a recent work \cite{rivera2023toward} proposed an MR interface for intuitively teaching trajectories to a holographic collaborative manipulator.  All of the aforementioned works, however, lack a homogeneous, structured representation of the underlying communication acts allowing operators to transfer skills to the robotic teammate. Additionally, they lack an empirical assessment of the demonstration capacities and perceived users' experience of these solutions. 

Therefore, unlike previous research, in the present article we aim at consistently framing KT inside the holographic communication space introduced in \cite{maccio2022mixed} and present a standalone approach for MR-based KT for any robot which can be described through the URDF format.  Furthermore, we provide an experimental evaluation of the communicative capabilities offered by our MR-based KT tool, assessing the learned robot skills in an interactive human-robot task. Finally, the proposed framework, adhering to the open-source paradigm, is made publicly available to other researchers and companies, who can employ it off-the-shelf as an alternative to traditional KT with any URDF-compatible robot, with minimum hardware setup required\footnote{\url{https://github.com/TheEngineRoom-UniGe/RICO-MR/tree/kt}}.

\section{Formalization}
\label{section:formalization}

\begin{figure*}[t!]
    \centering
    \includegraphics[width=0.9\textwidth]{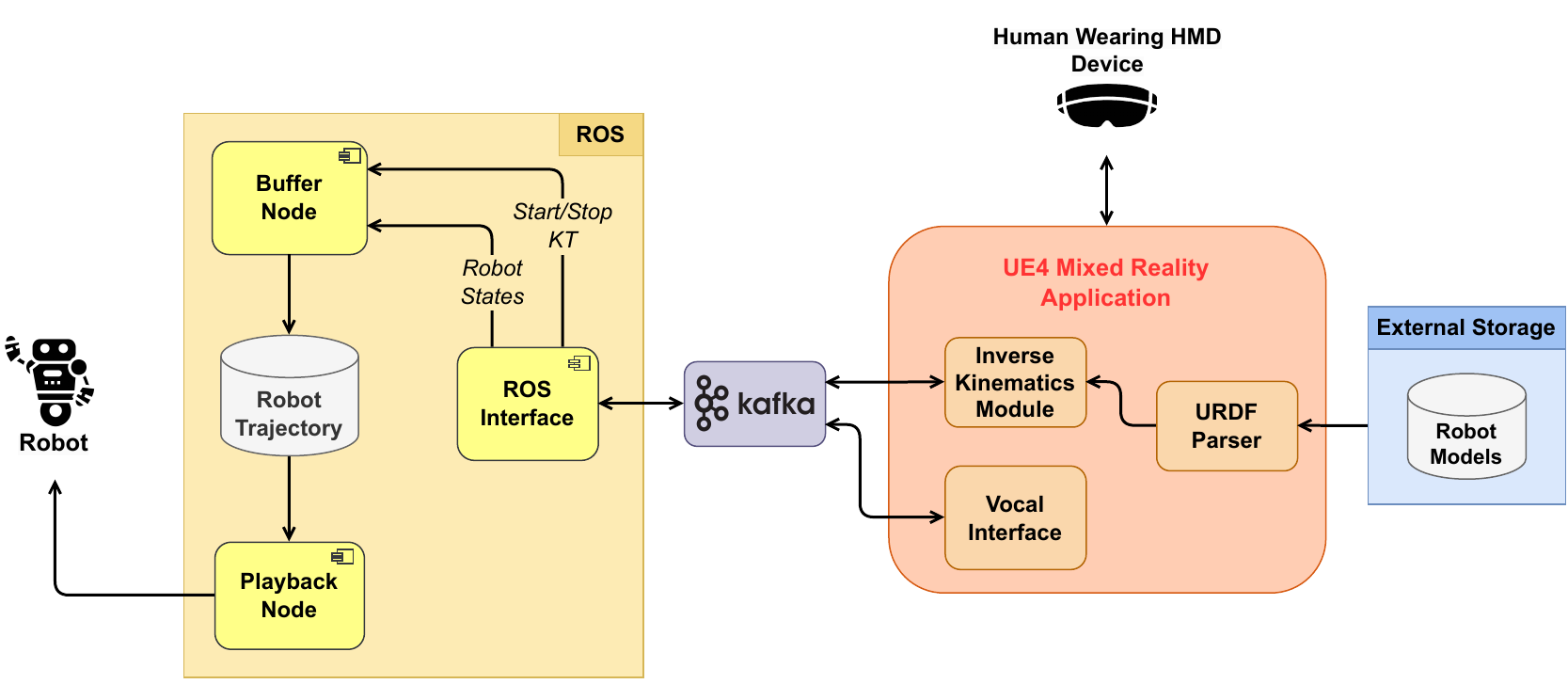}
    \caption{Overview of the proposed architecture implementing holographic KT, extending the framework detailed in \cite{macciò2023ricomr}.}
    \label{fig:software-architecture}
\end{figure*}

Recalling the definition provided in \cite{maccio2022mixed}, we describe communication as the act of conveying or transmitting \textit{pieces of information} (\textit{I}) through one or more communicative channels. It is noteworthy to mention that, in general, conveying a single piece of information may involve simultaneously multiple channels to strengthen the clarity of the communicative act itself. For example, human-human communication often combines verbal and gestural media to be meaningful and unambiguous. Following this principle, and denoting $M = \{m_1, \dots, m_{|M|}\}$ the set of all possible communicative media available (e.g., voice, gestures, gaze and so on), we provided the general formulation of a communicative act, namely

\begin{equation}
\label{eq:comm_act}
C(I, \bm{t}) = \bigcup_{i=1}^{N} C_{m_i} (I, \bm{t}_i) \, ,
\end{equation}

where $\bm{t}$ represents the time interval associated with the overall communication, whereas the intervals $\bm{t}_i$ span the duration of the individual components of the communication act.

Here, we leverage such formalization to frame KT inside the holographic communication space developed for \cite{maccio2022mixed}. The first step requires identifying the relevant information exchanged during KT sessions. In particular, we argue that the act of KT implies teaching robots about their future \textit{states}, denoted as $\bm{\tau}$. Without loss of generality, such a notion of robot state includes the robot's pose $\bm{x}(t)$ (that is, its position and orientation in the environment) and its joint configuration $\bm{q}(t)$. Consequently, we can formalize the robot's state as

\begin{equation}
\label{eq:robot-state}
\bm{\tau}(t) = \left\{ \bm{x}(t), \bm{q}(t) \right\} \, .
\end{equation}

This, in turn, provides us with a suitable representation of the set of information  ${I}$  which can be conveyed through KT, namely  ${I} = \left\{ \bm{\tau}(t) \right\}$ . Having defined the set $I$, we observe that KT is achieved by hand-guiding the robot's wrist or end-effector. According to our proposed formalism, this act involves a gesture-mediated communication  $C_\text{gest}$  that enables users to teach robots about their future states in a simple way and can be described as follows:

\begin{equation}
\label{eq:kt}
C_\text{gest}(I, \bm{t}_\text{gest}) = \bm{\mbox{T}}(\bm{t}_\text{gest}) \, ,
\end{equation}

where $\bm{\mbox{T}}(\bm{t}_\text{gest})$ describes the robot trajectory that is conveyed via gestural guidance during the interval $\bm{t}_\text{gest}$ spanning the KT session and is defined as

\begin{equation}
\label{eq:robot-state-trajectory}
\bm{\mbox{T}}(\bm{t}_\text{gest}) = \left\{ \bm{\tau}\left(t_\text{gest,\,s}\right), \dots, \bm{\tau}\left(t_\text{gest,\,e}\right) \right\} \, ,
\end{equation}

with $t_\text{gest,\,s}$ and $t_\text{gest,\,e}$ representing the temporal endpoints of the taught robot trajectory.

With this formalization in mind, we claim that KT can be translated and framed into the holographic communication space envisioned in \cite{maccio2022mixed} by letting users convey robots' trajectories via gestural guidance on a virtual counterpart of the robot.  As already mentioned, the unconstrained nature of the MR space allows for such a form of KT while solely relying on the built-in hand-tracking capabilities of the MR-HMD device. Additionally, such decoupling between physical and holographic layers could be particularly effective in production environments, as the operators could leverage the virtual robot to program or teach upcoming tasks, without halting the execution of real robotic chains. 

To further strengthen the communicative framework and ensure a more natural interaction, we postulate that adding the vocal medium would improve users' experience, enabling them to control more detailed aspects of the KT session, including the \textit{start} and \textit{stop} on the taught robot trajectory, or the possibility to \textit{open} and \textit{close} the robot's gripper for teaching pick-and-place actions. According to such modelling, the holographic-based KT process is translated into a communication act combining gestural and vocal interaction and, as such, can be formalized as follows:

\begin{equation}
\label{eq:holographic-kt}
C^{KT}(I, \bm{t}) = C_\text{gest}(I, \bm{t}_\text{gest}) \,\; \cup \,\; C_\text{voc}(I, \bm{t}_\text{voc}) \, .
\end{equation}

This formalization, combined with equation \eqref{eq:kt}, describes the building blocks of the communication act taking place during the proposed holographic-based KT process. In the following paragraph, these building blocks are translated into modular software components and integrated into a preexisting MR-based architecture.


\section{Software Architecture}
\label{section:softarchitecture}

The software components developed in the context of this work constitute a modular extension of the open-source architecture, named \textit{Robot Intent Communication through Mixed Reality} (RICO-MR), which is introduced and detailed in \cite{macciò2023ricomr}. The features described in this paragraph are publicly available under MIT licence in a separate branch of the main RICO-MR repository. A link to the repository is included at the end of Section \ref{section:background}.

The proposed architecture exploits functionalities developed for RICO-MR to achieve the holographic KT envisioned in Section \ref{section:formalization}. However, currently, the architecture allows holographic KT with fixed manipulators only. As such, we introduce a simplification in the formalization provided in \eqref{eq:robot-state}, and we hereafter refer to the notion of robot state to indicate its joint configuration $\bm{q}(t)$ only.

\subsection{Mixed Reality Application}

\begin{figure}[t!]
    \centering
    \includegraphics[width=0.44\textwidth]{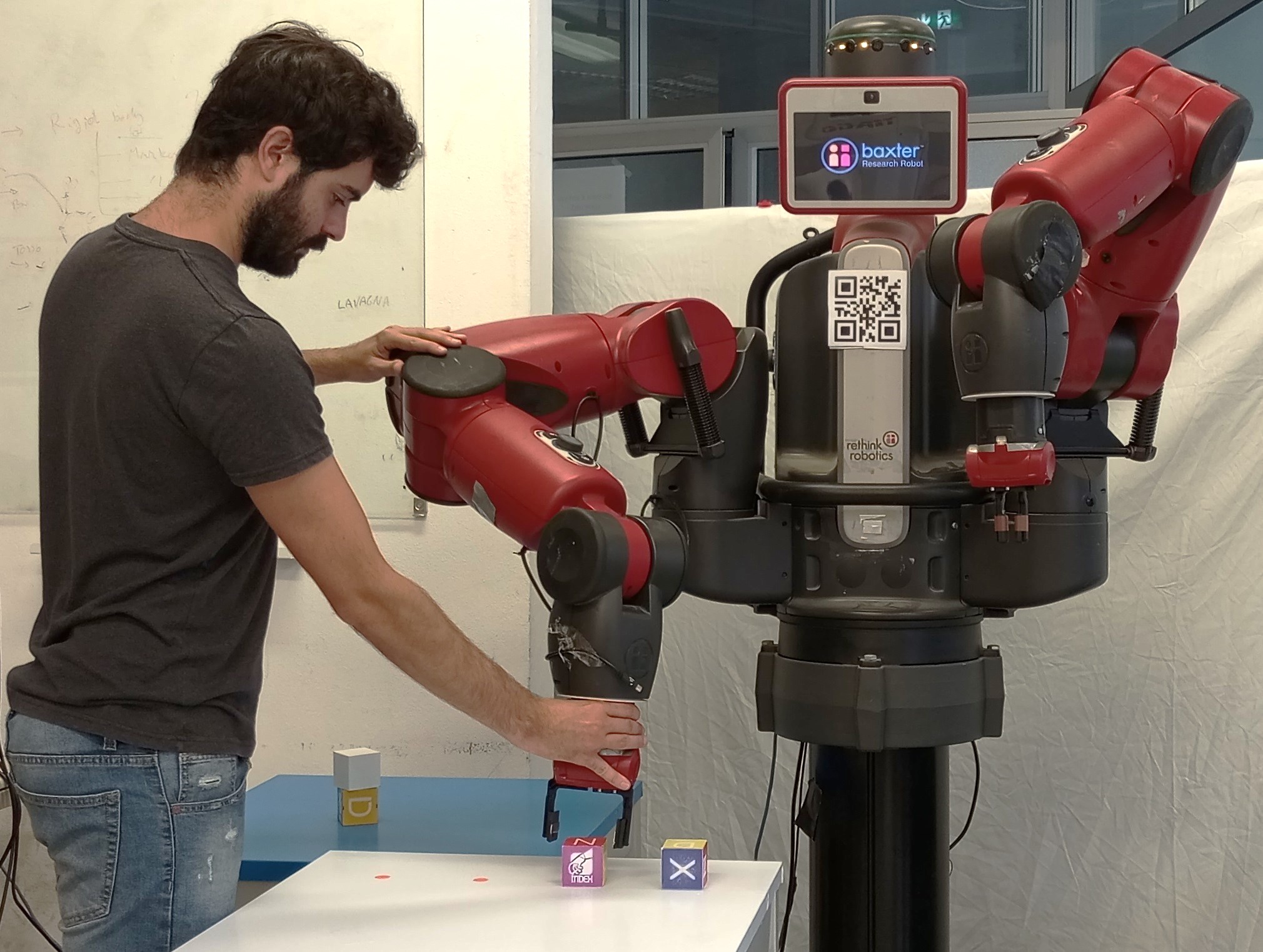}
    \caption{An experimenter interacting with Baxter during physical KT session. The operator drives the robot's arm through gestural interaction, teaching the sequence of pick-and-place actions needed to complete the stacking task.}
    \label{fig:baxter-kt}
\end{figure}

A \textit{MR Application}, built with Unreal Engine 4.27 (UE4) and deployed on the embedded HMD device worn by the user, drives the whole holographic interface. A hand-attached menu enables the user to select robot models from a list of predefined ones, making it possible to load and spawn holographic robots in the environment. Aside from the pre-loaded models that ship with the current architecture version, the list of supported robots can be extended by uploading relevant resources (i.e., URDF files) to a remote repository, which can be customized in the application's settings. As such, it is possible to employ the proposed application to carry out KT with any URDF-compliant robot.

Upon selecting the robot model, users can spawn it in the environment using a QR code as a spatial anchor, taking advantage of Unreal's marker detection capabilities. Along with the robot model, a grey holographic sphere, visible in Fig. \ref{fig:tiago-kt}, is spawned and superimposed on the robot's wrist. 
This sphere serves as a point of interaction between the human and the robot. Using the hand-tracking capabilities of the HMD, the human can directly manipulate the sphere by controlling its rotation and translation in space. The robot, in turn, follows the sphere and aligns its wrist's pose with it by solving the Inverse Kinematics (IK). To this extent, the Denavit-Hartenberg (DH) parameters necessary for the computation of the IK are extracted from the robot model's URDF and fed to the \textit{IK Module}, which continuously computes the joint configuration needed to achieve the desired pose of the wrist. Specifically, the IK computation occurs with a rate of $30\,Hz$. As such, by interacting with the grey sphere and hand-guiding it, users can communicate future robot's states and, consequently, teach trajectories and actions to the robot teammate. 

Consistently with the formalization given in Section \ref{section:formalization}, a voice interface is also active inside the MR application. Four basic commands are available, ensuring that the user can control the \textit{start}$\,/\,$\textit{stop} of the KT session and the \textit{open}$\,/\,$\textit{closed} state of the robot's gripper, offering the possibility to teach more complex motions such as pick-and-place or handover actions.

\subsection{Recording and Playback}

While the MR application provides the holographic interface to perform KT, recording and subsequent playback of the robot's actions are respectively managed through Apache Kafka and the Robot Operating System (ROS) \cite{quigley2009ros} framework. On the one hand, we take advantage of Kafka, an open-source, high-performant data streaming platform, for input$\,/\,$output data exchange with the MR application. Kakfa provides numerous advantages for real-time data streaming applications, including cloud integration and scalability, and it has been adopted for developing RICO-MR \cite{macciò2023ricomr}. In this context, we use Kafka to stream the robot's states at a rate of $20\,Hz$, beginning as soon as the user signals the start of the KT session through vocal command.

On the other hand, two ROS nodes act respectively as \textit{Buffer} for the robot trajectory streamed through Kafka and \textit{Playback} of the recorded motion. The \textit{Buffer Node} subscribes to the Kafka topic to access the robot's states, and it saves them to file for later execution. To this end, a \textit{ROS-Kafka Interface} has been developed to convert incoming Kafka messages into their equivalent ROS representation. Finally, the \textit{Playback Node} forwards state commands to the internal low-level controller of the robot at the same rate as the recording to reproduce the desired motion.

\section{Experimental Validation}
\label{section:experimental}

\subsection{Hypotheses and Experimental Scenario}

\begin{figure}[t!]
\centering
\begin{subfigure}{.45\textwidth}
\centering
\includegraphics[width=\linewidth]{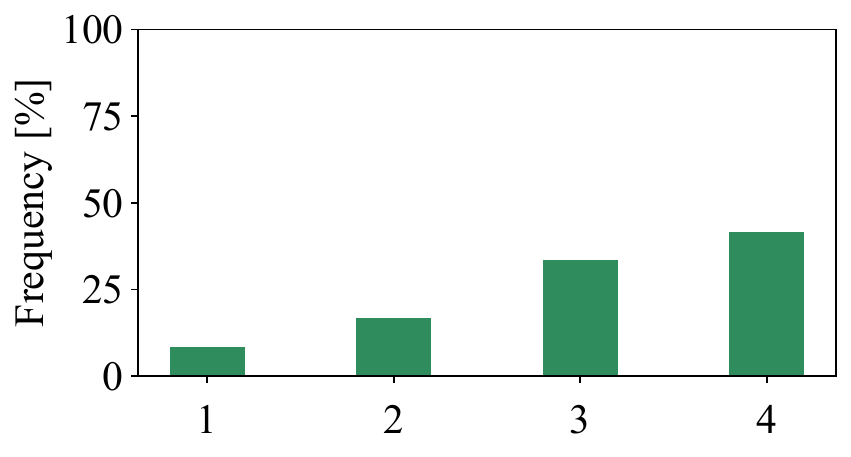}  
\caption{Cubes stacked in condition \textit{C1}.}
\label{fig:stacking-phys-kt}
\end{subfigure}
\begin{subfigure}{.45\textwidth}
\centering
\includegraphics[width=\linewidth]{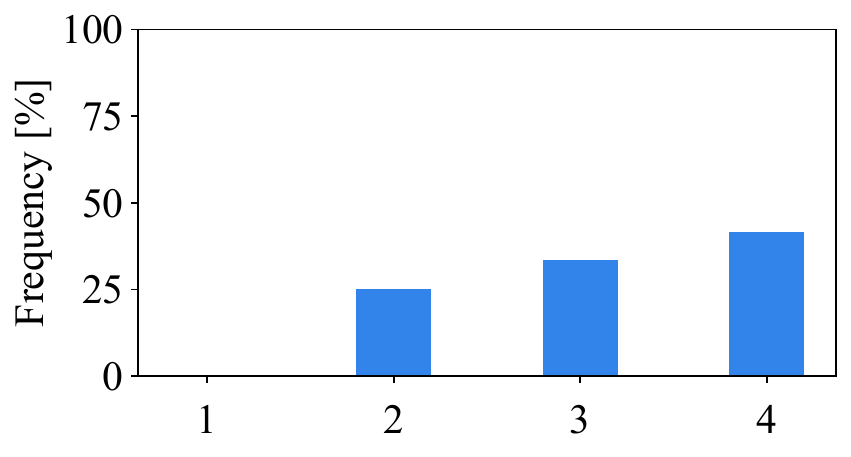}  
\caption{Cubes stacked in condition \textit{C2}.}
\label{fig:stacking-holo-kt}
\end{subfigure}
\caption{Histograms depicting the number of cubes successfully stacked by the robots during the playback phase, in the two experimental conditions.}
\label{fig:stacking-results}
\end{figure}

The experimental campaign carried out in this study aims to determine if our proposed holographic KT approach can act as a suitable alternative to standard, physical KT,  both in terms of demonstration capabilities and perceived user experience. 
To achieve our goal, we devised a human-robot interactive scenario to compare traditional physical kinematic teaching (KT), where the operator manually controls the robot's kinematic chain, with our proposed holographic approach. To ensure more generalized results, we conducted experiments using two different robots. In particular, we opted for \textit{Baxter} \cite{fitzgerald2013developing} from Rethink Robotics and \textit{Tiago++} \cite{pages2016tiago} from Pal Robotics, both being well-known platforms adopted in relevant research studies \cite{rosen2019communicating, maccio2022mixed, bongiovanni2022gestural, rosen2020mixed, ruffaldi2016third} and natively endowed with the necessary software and hardware components to achieve physical KT. Similarly, the HMD platform employed for rendering the holographic medium is a Microsoft HoloLens 2, a popular MR headset offering many features, including state-of-the-art hand tracking and voice interaction.

From a formal point of view, to provide a thorough comparison between physical KT and holographic KT, we have come up with the following hypotheses, which have been evaluated through preliminary user study: 

\begin{itemize}
    \item[\textit{H1}] There is no observable difference between actions taught through physical or holographic KT, namely the two approaches provide equivalent communicative power, leading to similar playback outcomes;
    \item[\textit{H2}] No difference can be observed in terms of temporal overhead when demonstrating actions through either physical or holographic KT;
    \item[\textit{H3}] No difference can be observed between the two approaches in terms of perceived UX during the demonstration process.
\end{itemize}

Regarding the interactive task employed to evaluate the two KT alternatives, a simple \textit{stacking task} has been devised. Specifically, the human should use KT to teach a sequence of pick-and-place actions aimed at stacking four cubes on top of each other according to a predefined order. Fig. \ref{fig:baxter-kt} depicts the experimental scenario, showing a user in the middle of a physical KT session with the Baxter robot.

\subsection{User Study}

We carried out a within-subject experimental campaign with $K=12$ volunteers (9 males and 3 females), all aged between 21-32 (\textit{Avg} $=26.3$, \textit{StdDev} $= 3.07$) and having limited or null experience with MR and HMD devices. The subjects were divided into two groups. The first group performed the experiment with Tiago++, while the second group used Baxter. In both groups, subjects were asked to perform the KT session in two different experimental conditions, namely
\begin{itemize}
    \item[\textit{C1}] Without wearing the HMD and performing physical, hand-guided KT.
    \item[\textit{C2}] Wearing the HMD and performing holographic KT.
\end{itemize}

\begin{figure}[t!]
    \centering
    \includegraphics[width=0.4\textwidth]{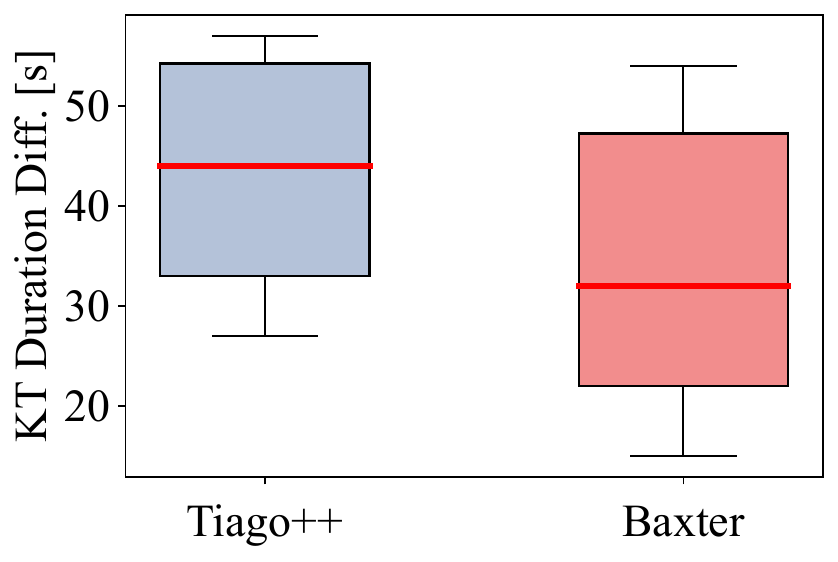}
    \caption{Differential distributions depicting the temporal overhead introduced by the MR medium when performing KT under \textit{C2}.}
    \label{fig:kt-time-diff}
\end{figure}

To avoid introducing unwanted biases, the starting experimental condition for each subject was randomized. Participants were initially instructed on the stacking task and assigned an arbitrary order for the cubes to be collected. Then, they performed their first trial, in condition \textit{C1} or \textit{C2}. However, before beginning the experiment with HMD on (i.e., condition \textit{C2}),  subjects were also briefly instructed on how to interact with the HoloLens holographic menus and interface. Then, once accustomed, they proceeded to carry out their trial.  Subsequently, each subject repeated the experiment in the opposite condition.  To achieve a consistent KT experience, the holographic interface in condition \textit{C2} also included four virtual cubes placed coherently with their real-world counterparts, as shown in Fig. \ref{fig:tiago-kt}. Such virtual cubes were physics-enabled and behaved like the real ones, aiding the participant in recording the holographic KT session. In both cases, the voice interface was active for controlling the \textit{start}$\,/\,$\textit{stop} of the KT session and the \textit{open}$\,/\,$\textit{closed} state of the robot's gripper. However, while in condition \textit{C2} the vocal interface was embedded into the MR application running on the HoloLens 2, in condition \textit{C1} it was simulated thanks to a \textit{Wizard of Oz} approach.

\begin{figure*}[t!]
\centering
\captionsetup[subfigure]{justification=centering}
\begin{subfigure}{.32\textwidth}
  \centering
  \includegraphics[width=\linewidth]{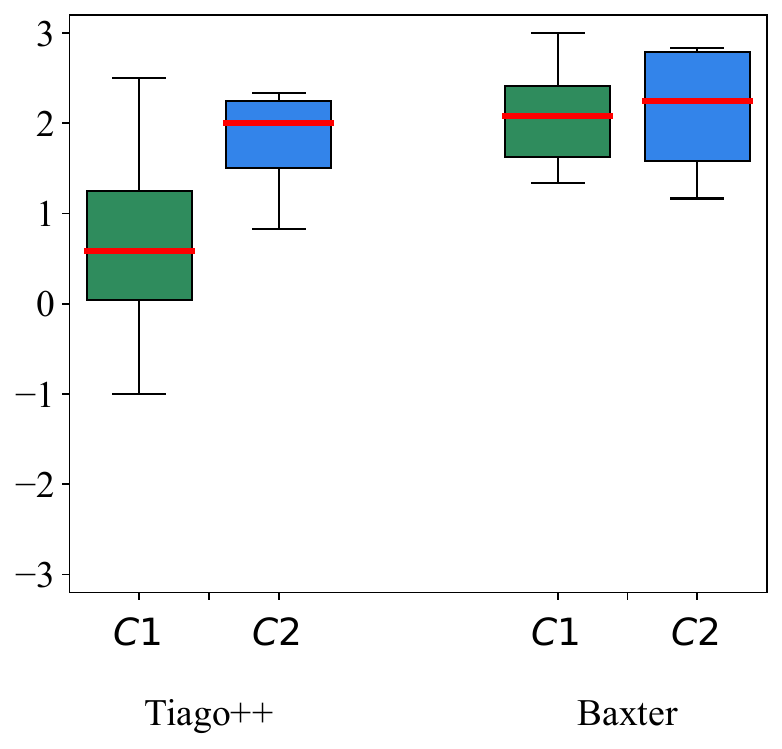}  
  \caption{Attractiveness}
  \label{fig:Attractiveness}
\end{subfigure}
\begin{subfigure}{.32\textwidth}
  \centering
  \includegraphics[width=\linewidth]{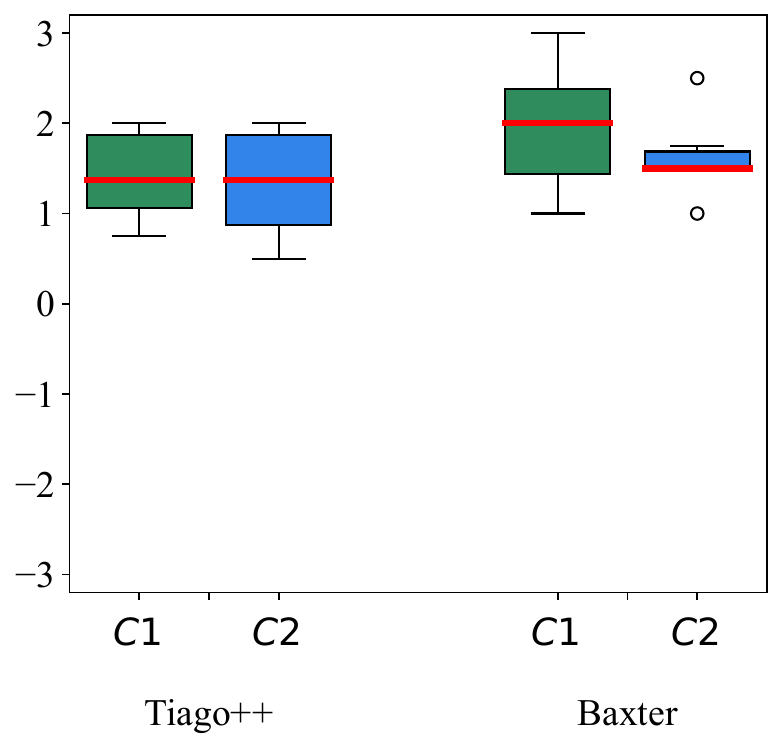}  
  \caption{Perspicuity}
  \label{fig:Perspicuity}
\end{subfigure}
\begin{subfigure}{.32\textwidth}
  \centering
  \includegraphics[width=\linewidth]{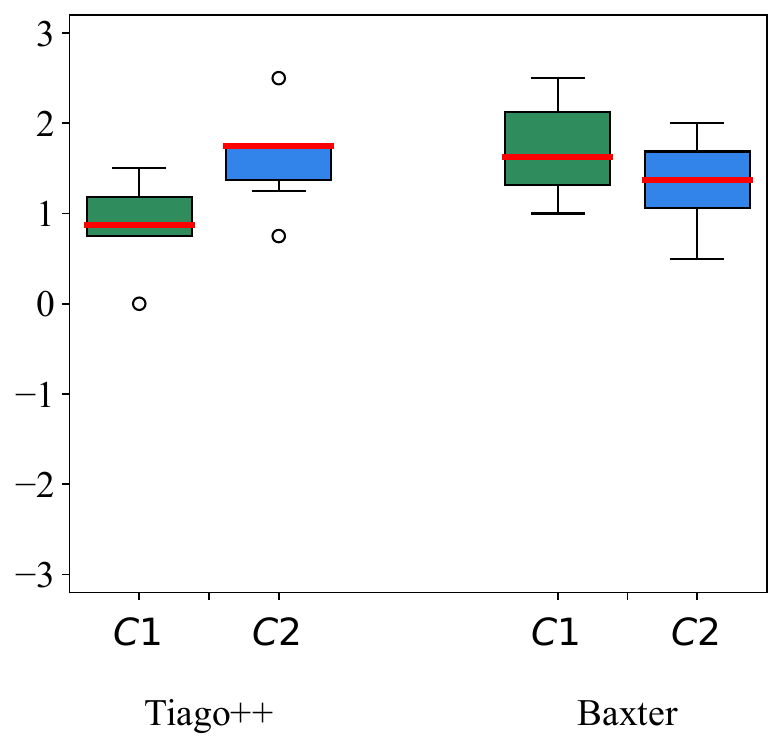}  
  \caption{Efficiency}
  \label{fig:Efficiency}
\end{subfigure}
\par\bigskip
\begin{subfigure}{.32\textwidth}
  \centering
  \includegraphics[width=\linewidth]{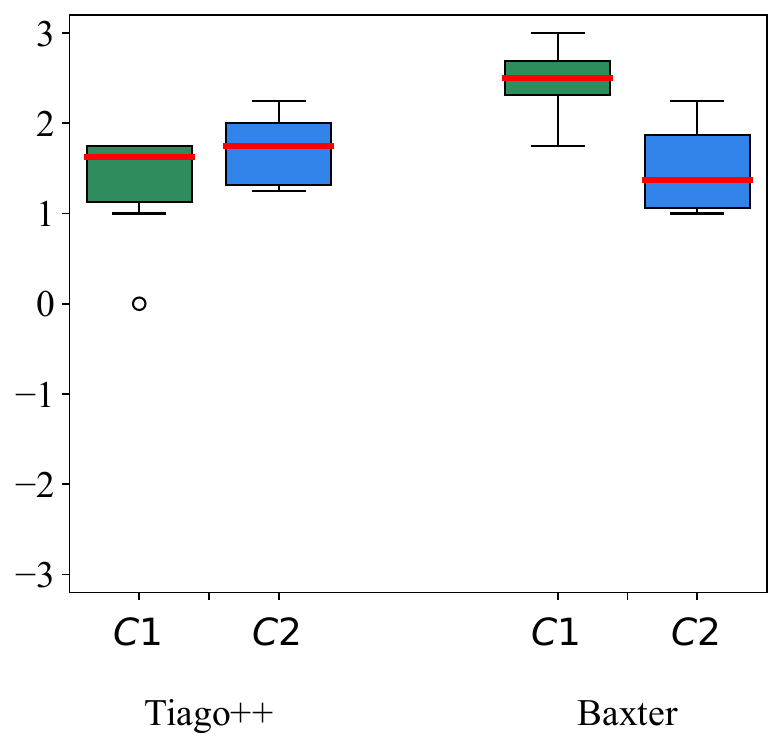}  
  \caption{Dependability}
  \label{fig:Dependability}
\end{subfigure}
\begin{subfigure}{.32\textwidth}
  \centering
  \includegraphics[width=\linewidth]{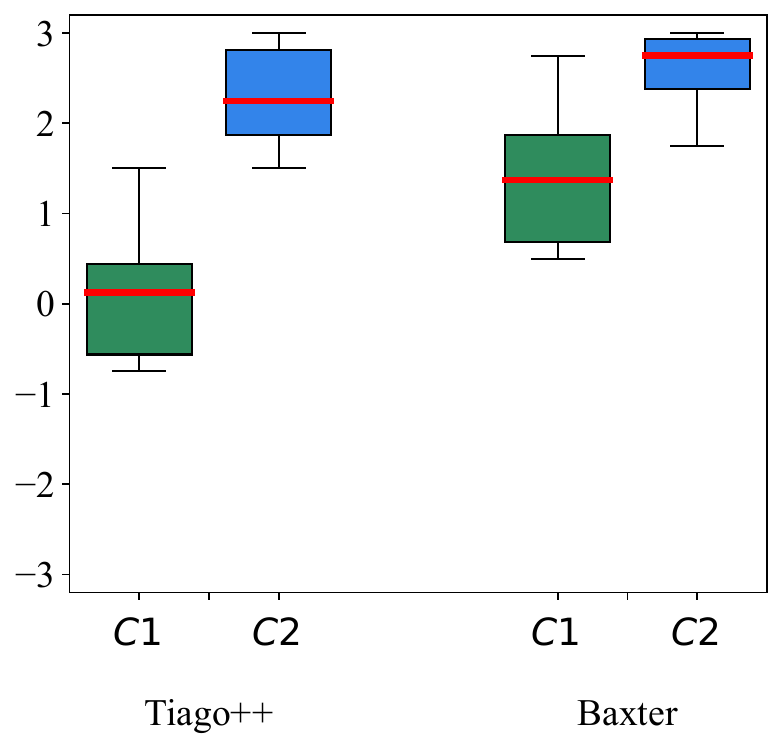}  
  \caption{Stimulation}
  \label{fig:Stimulation}
\end{subfigure}
\begin{subfigure}{.32\textwidth}
  \centering
  \includegraphics[width=\linewidth]{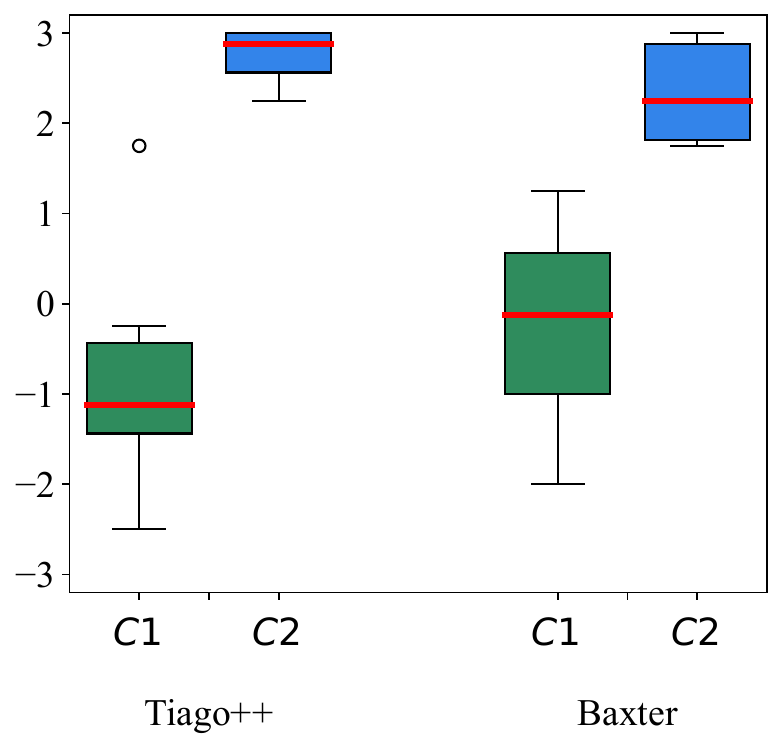}  
  \caption{Novelty}
  \label{fig:Novelty}
\end{subfigure}
\caption{Measured UEQ scores on the six evaluation scales, grouped by robot type and experimental conditions. The median value for each distribution is plotted as a red line.}
\label{fig:questionnaires-scores}
\end{figure*}

After successfully completing each KT session, the playback phase was manually triggered, causing the robot to reproduce the taught action. This phase allowed us to rank the KT session quantitatively by combining two distinct variables, useful in evaluating \textit{H1}  and \textit{H2}. On the one hand, we counted the number of cubes successfully stacked by the robot during playback. As such, we were able to evaluate the communicative capabilities of each KT alternative, assessing how well the combination of vocal and gestural interface translated into the corresponding robot action. On the other hand, we recorded the duration of each demonstration session and employed such quantity to compare the two KT techniques in terms of time necessary to teach the full stacking task.  

Finally, after completing their trials, each participant was required to fill out the User Experience Questionnaire (UEQ) \cite{schrepp2017construction}, a well-known survey useful for ranking and comparing interactive products. In particular, such a questionnaire allows grading the UX of a given product through six evaluation scales, namely \textit{attractiveness}, \textit{perspicuity}, \textit{efficiency}, \textit{dependability}, \textit{stimulation} and \textit{novelty}. In accordance with hypothesis \textit{H3}, to provide a consistent comparison between the two KT techniques, each participant compiled the UEQ twice, thus evaluating both physical and holographic KT sessions from a UX point of view.


\section{Results}
\label{scetion::results}

We hereby report and discuss the results obtained from our preliminary user study. In particular, we observed that, regardless of the robot, the two groups of subjects achieved comparable results when teaching the stacking task in both experimental conditions. As such, Fig. \ref{fig:stacking-results} reports only the aggregated results, comparing conditions \textit{C1} and \textit{C2} without discerning the interactions occurred with Tiago++ or Baxter. The histograms show the percentage of playback sessions where the robot successfully stacked a certain number of cubes. For example, in both experimental conditions, around 40\% of the subjects achieved a flawless KT, resulting in the robot successfully stacking all four cubes while replaying the taught trajectory.

By observing the plots of Fig. \ref{fig:stacking-results}, it is possible to note how physical and holographic KT yielded comparable results. Keeping into account that such distributions could not be assumed normal, we chose to perform a statistical evaluation of the two conditions via a non-parametric test, namely through a one-tailed Wilcoxon signed-rank test \cite{wilcoxon1992individual}. The test provided a statistic $W = 20$, with $p$-value$ > 0.3$. Such result was compared with the critical value $W_c$ obtained from the literature \cite{wilcoxon1947probability} by fixing the population size $K$ and the significance level $\alpha = 0.05$. As such, the corresponding critical value was $W_c = 17$. Observing the condition $W > W_c$, we could not reject the null hypothesis.
This result may indicate that our initial hypothesis \textit{H1} was correct, suggesting that the two communicative interfaces (i.e., physical and holographic) ensure consistent performances while executing KT.

Regarding the overall time needed to perform KT, we observed that in condition \textit{C2} participants were always slower because of their limited expertise with MR devices. As such, we chose to perform a differential analysis by computing, for each participant, the difference in terms of time taken to complete the KT session between condition \textit{C2} and \textit{C1}. These results are reported in Fig. \ref{fig:kt-time-diff}. The boxplots highlight that, on average, holographic KT lasted, respectively, for Tiago++ and Baxter, $44$ and $32$ seconds longer than the corresponding physical sessions. Compared with the average times measured to complete the physical KT sessions with the two robots, the MR-based approach introduced, respectively, a mean temporal overhead of $37\%$ and $33\%$. 
Statistically, this result is corroborated by a one-tailed t-test carried out on the original distributions, which yielded $p$-values $< 0.05$, therefore enabling us to reject the null hypothesis for \textit{H2}. Nevertheless, although these preliminary results suggest that the holographic demonstration process is slower than the physical one, we argue that the individuals' limited experience with MR devices played a major role in increasing the time taken to teach the stacking task. Consequently, further study could be undertaken with a more expert population to corroborate or revisit this finding.

Nonetheless, Fig. \ref{fig:kt-time-diff} shows no significant difference between temporal overheads when using one robot or the other. This result is also confirmed by a one-tailed t-test on the two differential distributions, which yielded a $p$-value$ > 0.2$. In other words, the overhead introduced by the MR medium was consistent among the two robots. 

Finally, Fig. \ref{fig:questionnaires-scores} reports the results obtained from the UEQ questionnaires, grouped per evaluation scale and robot type. Here, scores range in the interval $[-3, 3]$, with positive values indicating features that users appreciate given a particular interface. Specifically, Fig. \ref{fig:Efficiency} and \ref{fig:Perspicuity} highlight that both KT approaches provided comparable results in terms of \textit{efficiency} and \textit{perspicuity} (i.e., how intuitive and pragmatic the interface appeared to users), regardless of the robot employed. Such results are corroborated by statistical analysis performed through the Kruskal-Wallis test \cite{kruskal1952use}, a non-parametric ANOVA. The test yielded, for both scales, $p$-values $> 0.05$, indicating no significant difference between the distributions. Again, this result could suggest that the hypothesis \textit{H3} was correct, with both KT strategies leading to similar perceived UX. It is also worth mentioning that holographic KT scored particularly well in terms of \textit{attractiveness}, \textit{stimulation} and \textit{novelty}, suggesting that participants found the interaction with the holographic environment more engaging and original compared to the physical one. The only scale where holographic KT did a slightly worse job is \textit{dependability}, which measures how safe and predictable the users perceive a given interface. In this case, physical KT was still perceived as more predictable, particularly with the robot Baxter, compared to the MR-based approach, which nonetheless obtained positive scores with both robots.

\section{Conclusions}
\label{section:conclusions}

In this paper, we proposed a novel communicative interface based on MR to achieve KT with any  URDF-compatible robotic manipulator platform.  We built on top of our previous works and expanded our communicative framework \cite{maccio2022mixed} to account for holographic-based KT as a form of human-to-robot communication. Then, we presented a software architecture translating the formalization into a practical MR application running on embedded HMD devices. We compared holographic KT with standard, physical KT in a preliminary user study involving multiple subjects and two different robots. The results suggest that holographic KT behaves comparably to physical KT, achieving similar task-based performances and user experience.  This finding suggests that the proposed methodology could be adopted as a suitable alternative to physical KT in experimental and manufacturing scenarios, decoupling the demonstration process and enabling operators to program robot tasks in the MR space, without halting the production flow of the machine. 

In future works, we will evaluate whether these findings can be generalized by conducting user studies on a wider population, considering different robots, and more structured human-robot interaction scenarios where the individual is required to teach more complex tasks through holographic KT.




\bibliographystyle{IEEEtran}
\bibliography{bibliography}

\end{document}